\documentclass[journal,twoside]{IEEEtran}
\IEEEoverridecommandlockouts 
\usepackage{makecell}
\usepackage{amsmath}
\usepackage{multirow}
\usepackage{indentfirst}
\usepackage{verbatim}

\usepackage{graphicx}
\graphicspath{}

\usepackage{cite}

\ifCLASSINFOpdf
\else
\fi
\begin{document}
%
\title{A Review on Robot Manipulation Methods in Human-Robot Interactions}
%
%
%

\author{Haoxu~Zhang,~\IEEEmembership{Student~Member,~IEEE,}
        Parham~M.~Kebria,~\IEEEmembership{Member,~IEEE,}
        Shady~Mohamed,
        Samson~Yu,~\IEEEmembership{Member,~IEEE,}
        and~Saeid~Nahavandi,~\IEEEmembership{Fellow,~IEEE}
\thanks{H. Zhang, P. M. Kebria, and S. Mohamed are with the Institute of Intelligent Systems Research and Innovation (IISRI), Deakin University, Waurn Ponds, VIC 3216, Australia (s222117241@deakin.edu.au; parham.kebria@deakin.edu.au; shady.mohamed@deakin.edu.au)}
\thanks{S. Nahavandi is with the Institute for Intelligent Systems Research and Innovation (IISRI), Deakin University, Waurn Ponds, VIC 3216, Australia, and also with the Harvard Paulson School of Engineering and Applied Sciences, Harvard University, Allston, MA 02134 USA (saeid.nahavandi@deakin.edu.au).}
}

%
%

{This work has been submitted to the IEEE for possible publication. Copyright may be transferred without notice, after which this version may no longer be accessible.}

%



\maketitle

\begin{abstract}
Robot manipulation is an important part of human-robot interaction technology. However, traditional pre-programmed methods can only accomplish simple and repetitive tasks. To enable effective communication between robots and humans, and to predict and adapt to uncertain environments, this paper reviews recent autonomous and adaptive learning in robotic manipulation algorithms. It includes typical applications and challenges of human-robot interaction, fundamental tasks of robot manipulation and one of the most widely used formulations of robot manipulation, Markov Decision Process. Recent research focusing on robot manipulation is mainly based on Reinforcement Learning and Imitation Learning. This review paper shows the importance of Deep Reinforcement Learning, which plays an important role in manipulating robots to complete complex tasks in disturbed and unfamiliar environments. With the introduction of Imitation Learning, it is possible for robot manipulation to get rid of reward function design and achieve a simple, stable and supervised learning process. This paper reviews and compares the main features and popular algorithms for both Reinforcement Learning and Imitation Learning.
\end{abstract}


%
\IEEEpeerreviewmaketitle
\section{Introduction}
%
%
%
%
\IEEEPARstart{T}{he} concept of robotics was first proposed by Isaac Asimov in 1940s \cite{isaac3laws}.  It has been widely implemented in our life with the rapid development of robots, such as aerospace, agriculture, voice service system, industrial manufacturing, etc. With the increased use of robotic technology, academics have increasingly turned their attention to studying the interaction between human and robots, also referred to as Human-Robot Interaction, or HRI. HRI is usually considered to be a new or cutting-edge robotics field. The primary focus of HRI is on creating robots or robotic techniques that can interact with human in a variety of settings or carry out various activities based on human requirements \cite{hridefine}. Numerous applications of HRI are believed to enhance people's lives. Most HRI applications are now still in the research stage. However, this is expected to change quickly \cite{hridefine}. For example, HRI has been widely used in service robots, educational robots, healthcare and personal assistance areas \cite{kebria2021stable,kebria2020adaptive}, teleoperation systems \cite{kebria2018control}, and autonomous driving systems \cite{hridefine}. The applications of HRI systems include the Baxter industrial assembly line robot, which was designed in 2018 with robot emotional expression screen \cite{r4}, leader-follower remote operating devices such as T-HR3 robots, Tesla and Google autonomous vehicles, and Paro companion and therapy robot worked as assistive robots in eldercare since 2006 \cite{hridefine}. 

The most significant and challenging part of HRI research is robot manipulation. Control of robots in HRI is mainly focused on satisfying the task requirement and providing proper performance to human in specific scenarios based on given observations and feedback \cite{r3}. This research area has covered a wide range of topics, including learning specific manipulation techniques via human's demonstration, learning abstract descriptions of a manipulation task suitable for high-level planning, and identifying an object's function through interaction \cite{r3}. There have been several studies to investigate on different concepts of robot manipulation. Robot manipulation includes the following key concepts: physical system control, non-holonomic constraints and manipulation patterns, interactive perception and verification, hierarchical task decomposition and skill transfer, manipulation generalizing through objects, and new concepts and structures discovery \cite{r3}. The manipulation of the robot's physical system, the perception of the status of the external environment, and the control of the robot in unfamiliar contexts using prior knowledge or expert data are the principal areas of current research on human-robot interaction manipulation. 

Recent reviews on HRI robotic manipulation elements or tasks include the following fundamental aspects.  

Pre-programmed motion. Pre-programmed motion is a crucial ability with many uses in the manipulation area. It is described as the capacity to carry out specific motions for any load linked to the robot's mechanical interface, subject to constraint on the workspace, speed, acceleration, and load weight, with the necessary repeatability and accuracy. The main job of industrial robots is to carry out pre-programmed movements\cite{r5}. 

Compliant motion \cite{r5}. Pure motion control cannot ensure the safety and reliable behaviors of the manipulator, whereas compliant motion control comprises all control techniques that deal with tasks in which a robot interacts with its environment \cite{r6}. 

Pick and place motion. This includes both structured and unstructured pick and place actions. Structured pick and place actions are repetitive movements for identical objects. This can be achieved by pre-programming offline. Unstructured pick-and-place actions are tasks that are not constrained so that it can perform highly repetitive actions on the same objects. Examples include path planning, gripper mechanical manipulation for many different objects, different tasks based on environment and human commands, etc. \cite{r5, r30, r31}.  

Combined manipulation. This robot control aspect includes non-grasping manipulation and pick-and-place grasping manipulation with moving requirements. Such as remote-control surgery, teleoperated bomb disposal robot, articulated tandem industrial robot \cite{r5}, Shakey robot, etc. \cite{r7}. 

Whole-body manipulation. When manipulating items, a robot can employ all or most of its other degrees of freedom in addition to its hands or arms. This is referred to as whole-body manipulation \cite{r8}. This perspective regards control not only by hands, but also by other parts of the body. Such as holding a door with a hip or turning on a switch with an elbow \cite{r5}.   
\begin{figure}[ht]  
\centering 
    \includegraphics[scale=0.2]{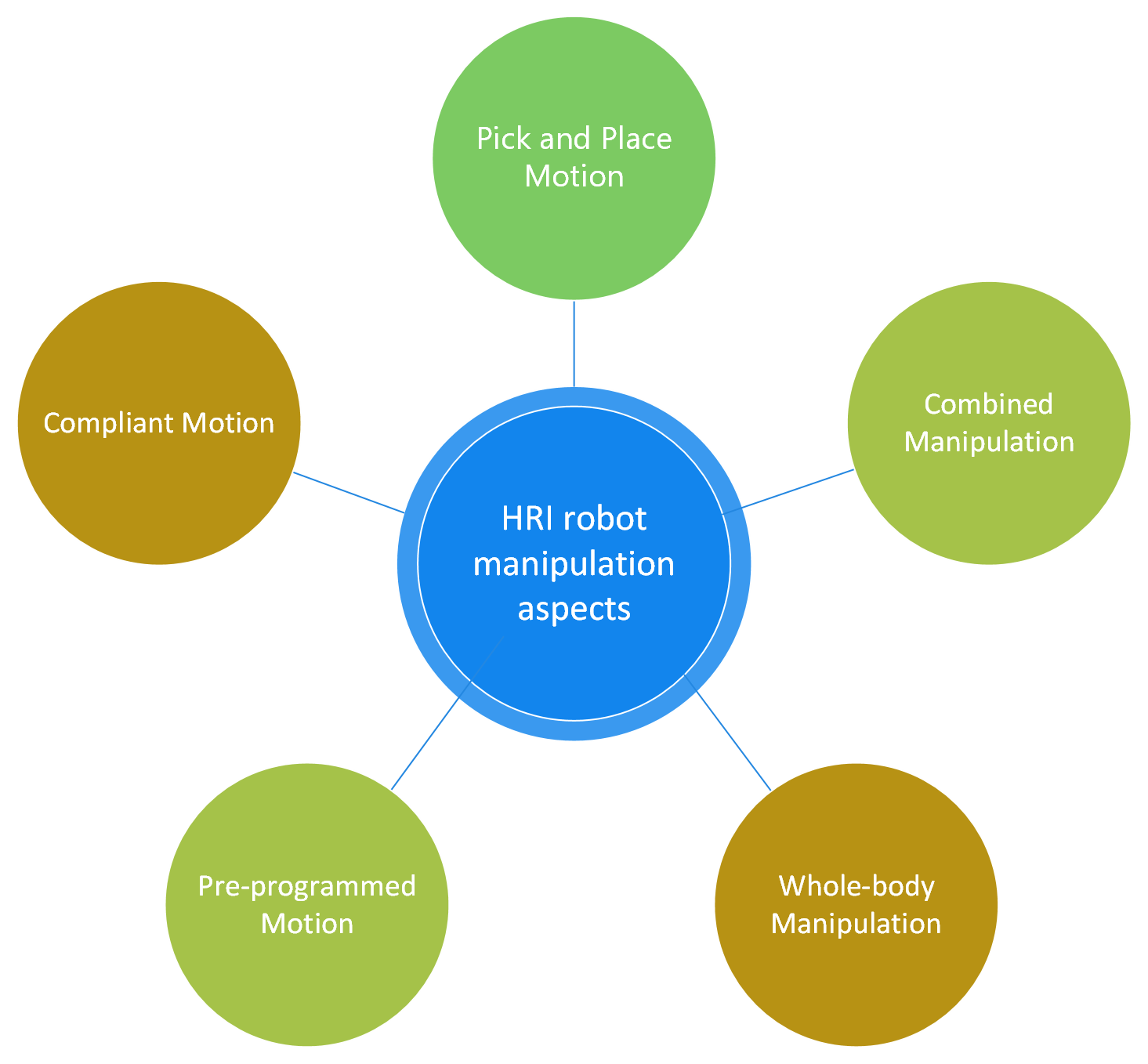} 
    \caption{Robot Manipulation aspects in HRI} 
    \label{Fig: Robot Manipulation aspects in HRI systems.} 
\end{figure}
 
All these human-robot interaction manipulation aspects discussed by previous literature require the robot to learn the manipulation of motor capabilities. Whereas robots learn to manipulate motor capabilities in addition to the traditional offline pre-programming approach where each action is defined in advance. Current cutting-edge research is ultimately aimed at helping a robot to learn the policy to achieve goals. However, skill learning policies for robot manipulation is hard as the following main factors. The mechanisms of the robots need to be feasible, wearable, easily fixed and affordable \cite{r5}. And the robot needs to handle a relatively comprehensive range of perceptions such as high-resolution pictures of object locations or applied force level \cite{r5}.  Apart from the above, robots also need to learn how to operate in dynamic and unknown environments \cite{r9}. The last one is how to simulate different training scenarios and build models to provide robots with enough training and upgrade the system furthermore.   

The literature studied robot learning policy for manipulation typically through machine learning algorithms with the ability of adaptive decision making and self-learning.  These robot learning algorithms can be divided into mathematical model-based methods and data-driven stochastic policies \cite{r3,r9}.  Researchers have developed many methodologies to deal with the challenges of HRI robot manipulation. The data-driven methods perform well in reducing computation complexity and handling uncertain tasks \cite{r9}. The deep learning network has reached outstanding achievements in machine perception and image processing attributed to its exceptional feature extraction capability. Reinforcement Learning \cite{r10} can be used to learn the policy parameters for skill controllers with given policy representation. Now with the development of Reinforcement Learning and its further expansions, for example, deep Reinforcement Learning, imitation learning \cite{r11}, etc., which make it possible for robots to interact with human to overcome challenges described in the previous paragraph.    

This review paper mainly provide a review of data-driven robot manipulation policy learning algorithms in human-robot interaction. The paper is organized as follows. Section 2 discusses the formulation of skill policy learning robot manipulation algorithm and types of state and action spaces. Section 3 presents state-of-art data-driven robot manipulation techniques proposed in recent years. In Section 4, the significant challenges and issues during the data-based control design process of HRI robot manipulation are discussed. Then following with the Conclusion in Section 5.  

\section{Robotic Manipulation Models}
Markov decision process (MDP) is one of the most popular approaches for modeling robotic manipulation, which contains state information as well as a transfer mechanism between states. This typical formulation addresses a variety of tasks, allowing researchers to create general-purpose learning algorithms that may be used for a variety of purposes \cite{r3}. To solve a practical problem with robotic manipulation, the first step is to abstract the practical problem into a Markov Decision Process. 

\subsection{Markov Process} 
Stochastic process is the” dynamics” part of probability theory. The object of probability theory is a static random phenomenon, while the object of stochastic process is a random phenomenon that evolves over time (e.g., weather changes or urban traffic changes over time). In a stochastic process, the value of a random phenomenon at a moment t is a vector random variable, denoted by $S_t$. All possible states form a set of states $S$. A stochastic phenomenon is a process of state change. The state $S_t$ at a moment $t$ usually depends on the state before moment $t$. We denote the probability that the state at the next moment is $S_{t+1}$ when the historical information $\left\{S_1, ..., S_t\right\}$ is known as $P\left(S_t+1|S_1, ..., S_t\right)$. 

A stochastic process is said to have the Markov property when and only when the state at a given moment relies primarily on the state at the previous moment, as indicated by the formula $P\left(S_{t+1}|S_t\right)=P\left(S_{t+1}|S_1, ..., S_t\right)$. That is, the current state is a sufficient statistic for the future, which indicates the previous state does not affect the current state that only the current state determines it. It should be evident that because a stochastic process exhibits the Markov property doesn’t indicate that it is entirely disconnected from history. Although the state at moment $t+1$  is only related to the state now, the current state still contains state information at moment $t-1$. Through this chain relationship, the information of history is passed to the present. Hence, Markov Process can considerably simplify the process when the current state is understood, and future can be predicted.  

The Markov process refers to a stochastic process with Markov properties, also known as Markov chain. This is usually described a tuple $\left<S, P\right>$, where $S$ is a number of state sets and P is a state transition matrix. Given a Markov process, it is possible to generate an episode from a state based on its state transfer matrix, a step also called sampling. 

\subsection{Markov Decision Process}
When an external "stimulus" causes this stochastic process to alter in concert, there is a Markov Decision Process (MDP). And the external stimulus in robotic control or HRI usually will be the action of agent which is the robot. Hence, Markov Decision Process is obtained by adding the action of agent into Markov Reward Process. The Markov Decision Process consists of $\left\{S, A, P, r, \gamma \right\}$ include:

\begin{itemize}
\item[$\bullet$] 
$S$ is the set of states. 
\end{itemize}

\begin{itemize}
\item[$\bullet$] 
$A$ is the set of actions. 
\end{itemize}

\begin{itemize}
\item[$\bullet$] 
$\gamma$ is the discount factor.
\end{itemize}

\begin{itemize}
\item[$\bullet$] 
$r\left(s,a\right)$ is the reward function, when the reward can depend on both the state and the action, and degenerates to $r\left(s\right)$ when the reward function depends only on the state. 
\end{itemize}

\begin{itemize}
\item[$\bullet$] 
 $P\left(s^, |s,a \right)$ is a state transfer function that represents the probability of reaching state $s^,$ after state $s$ performs action $a$.
\end{itemize}

Unlike Markov Reward Processes, in Markov Decision Processes, there is usually an agent that performs this action. For example, when a person randomly walking in a strange place is a Markov Reward Process, the person will receive a larger reward if he finds out the destination by luck; if there is a guidance helping him with the directions, the person can actively choose to go to the destination to receive a larger reward. The Markov Decision Process is a time-dependent ongoing process, and there are constant interactions between the intelligence and the environmental MDP \cite{r32}.

The control algorithm chooses the action $A_t$ in accordance with the current state $S_t$; the MDP then obtains $S_{t+1}$ and $R_t$ in accordance with the reward function and the state transfer function and feeds them back to the intelligence for use in further decision-making. A policy is a process by which the robot chooses an action from the list of possible actions $A$ depending on the present condition.

\subsection{Partially Observable Markov Decision Process}
Partially Observable Markov Decision Processes (POMDP) is the fundamental framework of robot manipulation in uncertainty tasks which combine hidden Markov models and normal Markov models that probabilistically relate unobservable system states to the observations \cite{r13,r33}]. The POMDP is a natural illustration of a sequential choice which issues uncertain action outcomes and a partially visible state. While the state is only partially observable, the action’s result is unknown \cite{r14}. Planning in nondeterministic and partially observable circumstances is applicable to many robotics challenges based on POMDP planning. For instance, in HRI systems the robot does not have further information on human intention and future movement so the human actions become uncertain for the robot \cite{r14}. 

In general, POMDP model is interpreted as a 6-tuple $\left<S, A, O, T, Z, R\right>$ with variables as below \cite{r14}:
\begin{itemize}
\item[$\bullet$] 
$S$ represents the state space which is the same as the normal Markov Decision Process. It includes the robot states and environment states. 
\end{itemize}

\begin{itemize}
\item[$\bullet$] 
$A$ is the action space with all actions the robot can perform. 
\end{itemize}

\begin{itemize}
\item[$\bullet$] 
$O$ indicates the set of all robot observations that the robot is able to observe. 
\end{itemize}

\begin{itemize}
    \item [$\bullet$] $T\left(s,a,s^, \right)$ represents transition function that affect actions uncertainty. 
\end{itemize}

\begin{itemize}
    \item [$\bullet$] $Z\left(s^,,a,o \right)$ is the observation function that indicates the noise measurement. This is a conditional probability function $P\left(o|s^,, a \right)$ which illustrates the post-observation after action $a$ in state $s^,$.
\end{itemize}

\begin{itemize}
\item[$\bullet$] 
$R$ is the reward function that could be parameterized by state-action pairs.  
\end{itemize}

Unlike MDPs, POMDPs allow agents to make state-dependent observations rather than directly observing the entire system state. The agent creates a belief about the state of the system based on these observations. A probability distribution of all potential states is used to express this belief, which is referred to as a belief state. A policy that details the activities to be taken in each belief state is the answer to a POMDP \cite{r13}. In the robotics aspect, POMDP model has been widely used in many areas such as machine navigation in underground mining, road condition prediction with weather uncertainty in autonomous driving, occlusion avoidance in industrial robot manipulation and object grasping, and limited communication in multi-robot cooperation tasks \cite{r33}. 

\section{Robotic Manipulation Algorithms in HRI}
\subsection{Robotic Manipulation}
Manipulation was described as "the function of utilizing the properties of a grasped object to fulfil a task" in a European research road map \cite{r5}. In the field of HRI, robot manipulation mainly refers to the manipulation of robots that interact with the environment.  In the fields of industry, logistics, healthcare, and other aspects, robotic manipulation is gaining ground \cite{r5}. From previous research, the widely used or popular topics in robotic manipulation include vision-based manipulation \cite{r17,r18,r20,r21}, contact-based robot manipulation [19] [22] [24], multi-robot manipulation \cite{r23,r25,r26,r27}, etc. Apart from the above, control methods of robot manipulation can be divided into two main aspects based on recent research, which are Reinforcement Learning and Imitation Learning.  Human behavior and sequential decision-making can be analyzed from a neuropsychological and cognitive science perspective via Reinforcement Learning (RL). Reinforcement Learning (RL) has been successfully applied in several dynamic robotic system control applications, including manipulation, locomotion, and autonomous driving, demonstrating the viability of the concept \cite{r28}. Instead of designing reward functions with trial and error collections for each task, Imitation Learning (IL) relies on expert demonstrations to emulate the desired behaviors \cite{r29}.    
\subsection{Reinforcement Learning}
While traditional robot manipulation learning methods require manually designed control strategies, the goal of modern AI development allows systems to learn autonomously and find optimal solutions. The development of modern AI allows systems to interact and exchange information with their surroundings without relying on a predefined control strategy. In addition, it is hard to continue to learn and update the choice of the optimal solution during operation. Reinforcement Learning is a branch of machine learning that allows agents to learn as they interact with their surroundings. Traditional RL is a principled mathematical framework for autonomous learning driven by experience \cite{P37}. Optimal policies are obtained based on an accurate model of the environment, but this is infeasible for real world robot manipulation problems. In addition, traditional RL is also limited by low-dimensional problems \cite{P37}. This limitation is due to the space complexity, computational complexity, and sampling complexity of RL as well as other algorithms \cite{r34}. 

\begin{figure}[ht]  
\centering 
    \includegraphics[width=\columnwidth]{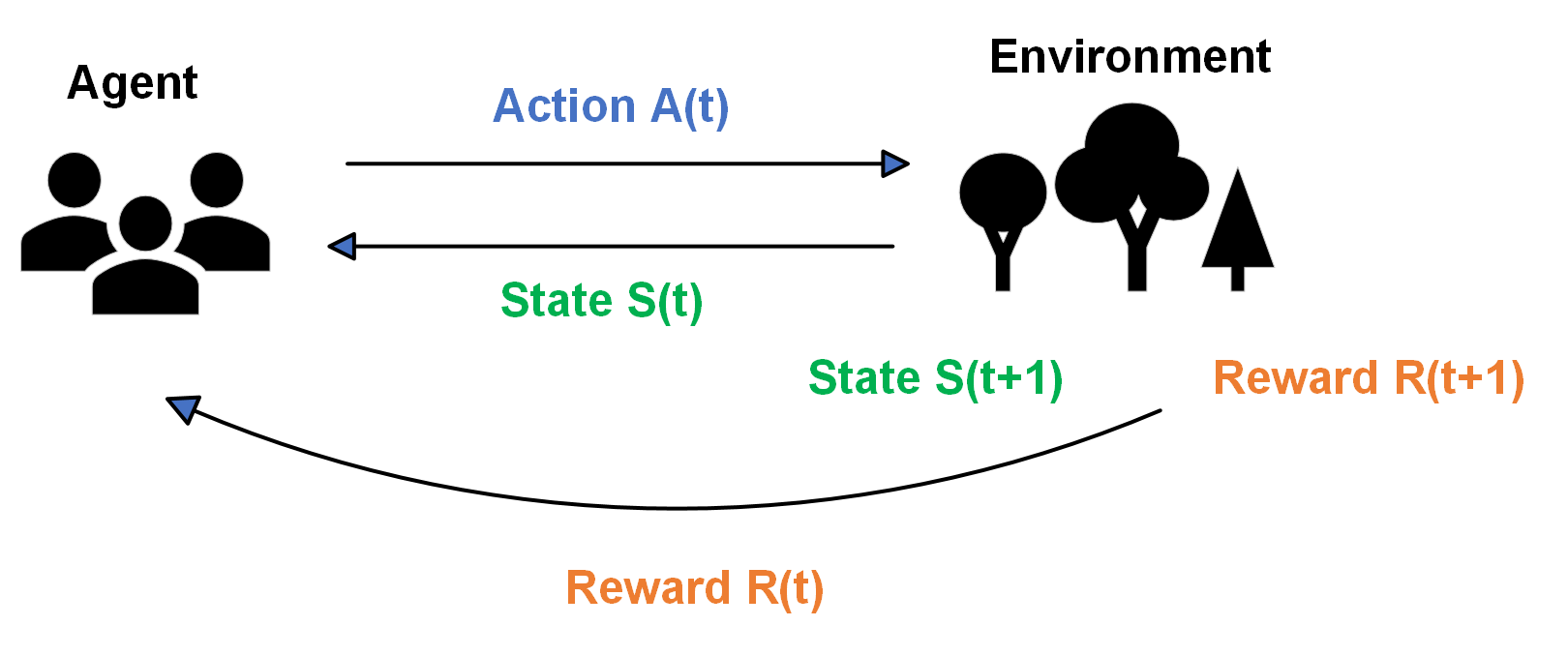} 
    \caption{Reinforcement Learning Process} 
    \label{Fig: Reinforcement Learning Process} 
\end{figure}

The fundamental goal of Reinforcement Learning is to seek the maximized expected return $G_t$ with discount factor $\gamma\in [0,1]$.
\begin{align}
G_t = r_t + \gamma r_{t+1} + \gamma^2 r_{t+2} + ...
\end{align}

In recent research, deep learning has received much attention because of its powerful function estimation, high-dimensional information processing, and representative learning capabilities. Many studies have shown that deep Reinforcement Learning (DRL) can be used to address the limitations of traditional Reinforcement Learning for robot manipulation, such as processing high-dimensional input signals and handling continuous state space tasks. For now, robots could learn behaviors directly from high-dimensional input signals like images with deep Reinforcement Learning, which combines the perceptual and decision-making abilities of deep neural networks \cite{robotcontrol}. Reinforcement Learning learns through interaction with the environment and can be tuned based on reward-driven behavior. In many current studies, popular deep learning methods that combine with RL include deep convolutional networks, deep autoencoders and deep recurrent networks \cite{P38}. 

There are two prominent success stories of deep RL in previous research. The first was at the beginning of DRL development, where it was experimented with Atari games with 2600 video games by learning directly from pixels. And further in Mnih, Jaderberg and Mirowski \cite{r35}, this illustrates that DRL can be trained with raw, high-dimensional observation. The second is the defeat of a human chess player by AlphaGo, trained by Reinforcement Learning and traditional heuristic search algorithms \cite{P37.128}. DRL has been applied to many fields, such as robotics, meta learn, and creating systems that can learn how to adapt in the real world.  
\\
\subsubsection{Reinforcement Learning Algorithms}
There are two primary classification methods for Reinforcement Learning, which have been used in numerous research projects. The first is determined by the knowledge of the state transition probability function and return function. This method mainly distinguishes whether RL requires an environmental model. In this regard, RL can be classified as model-free and model-based \cite{P37, robotcontrol}. The second type is based on the classification of deep learning methods applied in conjunction with RL. These can be classified as RL and Supervised deep learning combined methods, RL and Unsupervised deep learning combined methods, and Partially Observable MDPs (POMDPs) Environments based DRL \cite{P38}. In most reviews, DRL is classified according to the first method, and this section also follows this approach to review various algorithms for RL. 

\begin{figure}[ht]  
\centering 
    \includegraphics[width=\columnwidth]{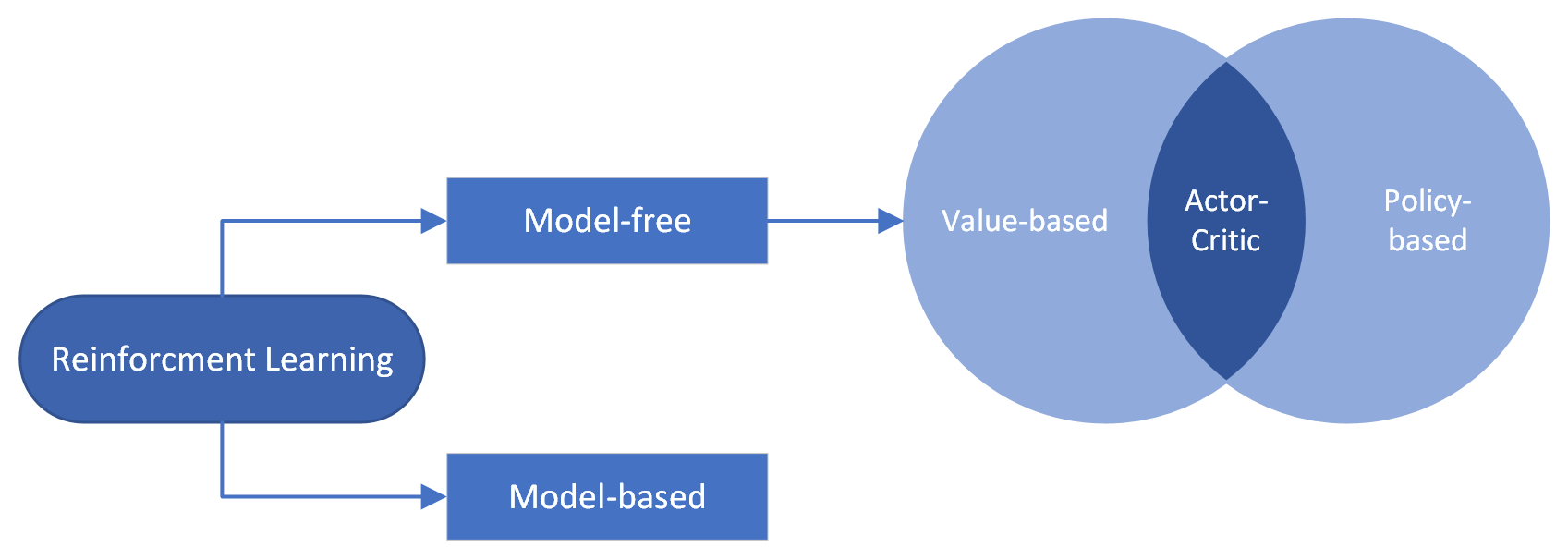} 
    \caption{Reinforcement Learning types} 
    \label{Fig: Reinforcement Learning types} 
\end{figure}

\paragraph{Model-free RL}
The model-free RL approach is to optimize the robot manipulation strategy through trial-and-error interactions between the agent and environment. Currently, there are three approaches of model-free Reinforcement Learning, which are based on value function, policy search, and a combination of both, actor-critic approach.

The value function in value-based RL normally refers to the accumulation of the expected return of the given state. Value functions can be roughly classified into the following three types, state-value function, state-action value/quality value, and learning the advantage function. 
The state-value function $V$ is a method to evaluate the absolute state-action values which is the expected return calculated when starting from state $s$ and subsequently following policy $\pi$. The optimal policy $\pi$ is defined as the policy with maximum state-value based on the state $s$ \cite{P37}.

Another commonly used value function is state-action value function (quality function) $Q$ which is the returned expected reward given with state $s$, initial action $a_0$ and policy $\pi$ followed only from subsequent states \cite{P37}.  
The value functions $V^{\pi}(s)$ of state s and $Q^{\pi}(s,a)$ of action-state can be presented as below. 
\begin{align}
V^\pi(s)=E_{a \sim \pi}[R(\tau)|s_t=s]
\end{align}
\begin{align}
Q^\pi(s,a)=E_{a \sim \pi}[R(\tau)|s_t=s, a_t=a]
\end{align}
With the optimal policy $\pi^*$ applied then optimal value functions can be presented as follow. 
\begin{align}
V^*(s)=\max\limits_{\pi^*}E_{a \sim \pi^*}[R(\tau)|s_0=s]
\end{align}
\begin{align}
Q^*(s,a)=\max\limits_{\pi^*}E_{a \sim \pi^*}[R(\tau)|s_0=s, a_0=a]
\end{align}

Apart from this, learning the advantage function $A$ is also implemented in value-based RL method. The advantage function denotes the relative state-action values compared with state-value which is absolute state-action values. It is defined as the relative advantage of actions based on $A^\pi=Q^\pi-V^\pi$ \cite{P37, p37.6}.  

There are some widely used and representative value-based RL algorithms, such as Q-learning and deep Q-network. The $Q$ value represents the expected reward in state $s$ when take action $a$ which is the state-action value introduced above. The primary idea of the algorithm is to establish a Q-table of state and action to store the Q-value, and then choose the activity that may obtain the highest reward according to the Q-value. The environment will give feedback on the relevant return reward $r$ according to the agent's action. Finding the expectation of the strategy with the biggest cumulative reward value is the aim of Q-learning. For example, using the time-difference updating approach, Q-table is updated by computing and learning the Q-value, where $\alpha$ is the learning rate and $\gamma$ is the rewarding discount factor.
\begin{align}
    Q(s_t,a_t)+\alpha \left[R_t+ \gamma \max \limits_a Q(s_{t+1},a)-Q(s_t,a_t) \right]
\end{align}
Q-Learning can be treated as value iteration, which means Q-Learning estimate the optimal $Q$ value $Q^*$ directly based on Bellman equation as follow. 
\begin{align}
    Q^* (s,a) = r(s,a) + \gamma \sum_{s^\prime \in S} P(s^\prime | s,a) \max\limits_{a^\prime} Q^* (s^\prime, a^\prime)
\end{align}
And Q-Learning updates are not necessarily to use the current value data sampled by $\arg\!\max\limits_a Q(s,a)$, which is because any $(s,a,r,s^\prime)$ can be used to update Q value.

However, Q-table calculation is quite expensive \cite{robotcontrol}. Hence Deep Q-network is proposed to explore high-dimensional space which combines Q-learning and deep neural network. DQN and Q-learning are both value-based iterative RL algorithms, but in ordinary Q-learning, the Q-value of each state action pair is stored in a Q-Table when the state and action space is discrete and not high dimensional, when the state and action space is high dimensional continuous, it is very challenging to use a Q-Table because the state and action space is too large. Therefore, by fitting a function purpose to create Q-values rather of a Q-table, the Q-table update can be changed into a function fitting issue where similar states receive similar output actions\cite{link}.   

Value-based Q-learning also performs well in multi-agent reinforcement learning. With the development of DRL, multi-agent RL started to move from tabular methods to deep learning with the ability to handle high-dimensional state action spaces. The development of Q learning in multi-agent RL ranged from independent Q learning \cite{r38} to tabular Q learning, such as sparse cooperative Q learning \cite{r39}, and the development of value decomposition networks (VDNs) that can decentralize the execution of centralized value function learning \cite{r40}. \cite{r41} proposed an innovative approach QMIX with rich action-value functions that can be decentralized and trained in a centralized end-to-end manner. 

In addition to this, Q value can also be used in memory based DRL, which is built on several memory modules and derived from studies on episodic control of human brain episodic memory. Memory based RL is aiming to incorporate external memory to solve the high magnitude environment interaction challenges in robot manipulation in HRI aspect \cite{r36}. The application of memory is also introduced in Q learning based Meta Reinforcement Learning in \cite{r37}. With the combination of experience learned from similar tasks, Meta Reinforcement Learning demonstrates the efficiency of improving task performances \cite{r37}.  

The Policy-based model-free RL doesn't need to preserve the value function model, instead this method searches the optimal policy directly \cite{P37}. It is hard for value-based RL to ensure every action quality in continuous action so that optimal value is difficult to locate \cite{robotcontrol}. Value function might be needed in policy-based methods still; however, the final optimal policy is determined through value function parameters \cite{p11.61}. Policy-based RL is typically categorized based on gradient-based which is policy gradient-based RL and gradient-free methods which is evolutionary algorithm (EA).  

Stochastic policies back-propagation is a typical gradient-based RL method which allows the policy to learn stochastic strategies in a task-related way. Stochastic policies back-propagation is also a key concept in algorithms such as Stochastic Value Gradient (SVG) \cite{p37.46}. For example, when deciding where to look in an image to track, classify, or illustrate objects, random variables will determine the coordinates of a small portion of the image, thus reducing the amount of computation required. While directly searching for the optimal policy can be hard and limited by local minima \cite{P37}, trust region policy optimization (TRPO) is proposed to update the policy parameters through surrogate objective function optimization \cite{robotcontrol, p37.123}. This method has shown that it is robust and useful especially with high-dimensional inputs\cite{p37.122} Later algorithm method proximal policy optimization (PPO) obtains better performance than TRPO in robot manipulation in virtual simulation environment \cite{robotcontrol}. This method only needs first-order gradient information.  

Evolutionary algorithm (EA) is typically defined as a search algorithm that consists of new solution generation, solution adjustment, and solution selection. Each element of the evolutionary algorithm in this work defines a deep neural network. The weights (genes) of these neural networks undergo random perturbations through mutation. Evolutionary algorithms can be closely related to neural networks evolving which is also called neuroevolution \cite{r42}. The combination RL algorithm ERL of gradient-based RL and EA demonstrated noticeably better performance than earlier DRL and EA techniques on a variety of difficult continuous control benchmarks \cite{r42}.  

Policy-based methods get rid of large amount of value computations, but the optimal policy can only be updated at the end of the round \cite{robotcontrol}. 

Consequently, actor-critic approach is proposed that combines both value-based and policy-based methods advantages. Actor-Critic consists of two parts by name, Actor and Critic which also means the combination of actor-based only algorithm (value-based) and critic-based only algorithm (policy-based). What the Actor does is interacting with the environment and learning a better strategy using a strategy gradient guided by the Critic value function. What Critic does is learning a value function from the data collected by the Actor interacting with the environment. This value function is used to determine whether the actions are good or not in the current state, which in turn helps the Actor make policy updates. 

Actor-critic algorithms can produce continuous actions like value-based RL. In addition, the role of the critic is evaluating the current prescribed action by the actor \cite{p39}. The assessment can be performed by following any policy evaluation principles, for example, Temporal Difference (TD), Least Squared Temporal Difference (LSTD) or residual gradients. After action evaluation by critic, the value function is updated by critics using samples. Then the updated value function is implemented on the actor's policy to further improve the updated policy's performance \cite{p39}.  

In the past research, actor-critic RL approach has been extended from learning simulated tasks to learning real robot vision navigation tasks directly from image pixels. For example, deterministic policy gradients (DPGs) extend the standard strategy gradient theorem from stochastic to deterministic strategies \cite{P37}. The main advantage of DPG is that DPG is integrated only on the state space and requires less samples in large action space states \cite{P37}. Apart from this, Deep deterministic policy gradients can be learned with continuous and low-dimensional observation tasks by using the same hyper-parameters and network structure \cite{p40}. Previous research demonstrated that DDPG can learn good policies from pixels directly with constant hyper-parameter and network structure \cite{p40}. However, DDPG is also facing overestimation problems when critic evaluate the Q value. In \cite{r43}, it proposed Twin Delayed Deep Deterministic policy gradient algorithm based on Double DQN to limit the overestimation issue in DDPG. In order to bootstrap off similar state-action pairs, it applies a regularization technique in the SARSA style, which alters the temporal difference target. It is proved that TD3 significantly improves performance and learning speed of DDPG \cite{r43}.  

Apart from overestimation problem, DDPG also requires numerous training sets to find the solution. Another representative actor-critic DRL is Asynchronous Advantage Actor Critic (A3C). A3C has achieved many state-of-art performances on many gaming tasks such as Atari 2600 to optimize policy and value function with single DNN \cite{p41}. In \cite{p41}, there is a parameter center to store the network weights which is like other asynchronous algorithms. The actor calculates gradients and updates to central server every few actions or terminate until reached the state. And the server propagates the new weights to actors to share common policy \cite{p41}. A3C methods perform well in high-dimension actions with easy data obtaining \cite{robotcontrol}. Recent studies introduced soft actor-critic (SAC) algorithm which is more efficient and stable for real-world robot control. The soft actor-critic method is proposed based on maximum entropy framework \cite{p42}. In \cite{p42}, the introduced soft actor-critic algorithm consists of three key parts: actor-critic structure with separate policy and value functions, an off-policy formulation to reuse previous collected data, and entropy maximization stability and future exploration\cite{p42}. For other model-free RL methods such as PPO, TRPO or A3C which all require collecting new samples nearly updates which is quite expensive on computation \cite{p42}, while soft actor-critic can perform robotic tasks in a short time and work in different environments with same hyper-parameters \cite{robotcontrol}. 

\paragraph{Model-based RL}
Despite model-free methods, there is another type of RL which is the model-based RL algorithm. Model learning methods can generate an environment model based on sample data or with the model provided. In real tasks, model learning could be limited due to the difficulty of obtaining an accurate environment model which could be time-consuming and the expensive computation burden. When the model is given, the optimal solution can be obtained by the value iteration algorithm and the policy iteration algorithm \cite{robotcontrol}. The value function approach allows for timely adjustment of the strategy based on the state values thus significantly reducing the iteration time \cite{robotcontrol}.  

As directly searching the optimal policy is hard and limited by local minima \cite{P37}. Hence, guided policy search (GPS) is proposed to learn from a few actions of other controllers and combine with important samples for off-policy sample corrections \cite{P37}. This method works in loop to match the sampled trajectories which is a combination of conventional control algorithm. GPS effectively balances the search from a good local optimum \cite{P37} but it is limited by the low efficiency of conventional control method. To overcome the limits of GPS, cross-entropy method is proposed to gain the probability of robot actions by taking samples and improve the efficiency. This is a simple algorithm to train agents which can also be used as a baseline. 

I2A is another model-based RL method for imaginative augmentation proposed by Deepmind, a new architecture for DRL that combines model-free and model-based. I2A makes the use of environmental models to enable the use of imagination to enhance the model-free agent. Given current information, future predictions can be made by querying the environmental model. Compared to most existing model-based RL learning and planning methods, I2A learns to interpret predictions from models of the learning environment, constructs implicit plans in an arbitrary manner, and uses the predictions as additional context for deep policy networks. I2A shows better data efficiency, performance, and robustness to model error specification \cite{r44}.  

Model-Based DRL with Model-Free Fine-Tuning (MBMF) combines neural networks with model predictive control (MPC) to propose an efficient model-based algorithm for solving high-dimensional, rich-contact problems, in which the model is built using deep neural networks, in addition to a method that uses a model-based learner to initialize a model-free learner to achieve high reward and significantly reduce sample complexity \cite{r45}. 

MVE introduces the environment model into the model-free RL method, i.e., Model-Based Value Expansion rolls out the environment model a certain number of steps before performing Q-value estimation. So that the target Q-value is simulated by the environment model first and then the Q-value is estimated. In this way, the Q-value prediction combines the short-term prediction based on the environment model and the long-term prediction based on the target Q network \cite{r46}.

\begin{table*}[!ht] 
    \centering
    \caption{Reinforcement Learning Algorithms compare - Model free}
    \label{Reinforcement Learning Algorithms compare - Model free}
    \resizebox{\textwidth}{0.13\textheight}{\begin{tabular}{|l|l|l|l|}
    \hline
        \makecell[l]{Method} & Category  & Advantages & Disadvantages \\ \hline
        \makecell[l]{DQN \cite{r49}} &  \makecell[l] {Value-based}  &  \makecell[l]{the applicability to scenarios with large state \\space and high dimensional space} &  \makecell[l]{poor stability and hard to output continuous policies} \\ \hline
        
         \makecell[l]{TRPO\cite{robotcontrol, p37.123}} &  \makecell[l]{Policy-based} &  \makecell[l]{robust and useful with high-dimensional inputs with\\ the control of step length that can achieve better assurance of \\monotonic improvement in each step} &  \makecell[l]{an excessively long step could result in abrupt \\and severe worsening of the policy that influences\\ the training effect} \\ \hline
         
         \makecell[l]{A3C \cite{p41}} &  \makecell[l]{Both} &  \makecell[l]{alleviates the difficulty of convergence of Actor-Critic and provides\\ a general asynchronous and concurrent RL framework} &  \makecell[l]{the obtained solutions are usually local optimal\\ evaluation strategies are usually not very efficient\\ and have a high bias}\\ \hline
         
         \makecell[l]{DDPG \cite{p40}} &  \makecell[l]{Both} &  \makecell[l]{can be learned with continuous and low-dimensional\\ observation tasks by using the same hyper-parameters\\ and network structure}& \makecell[l]{facing overestimation problems when critic\\ evaluate the Q value} \\ \hline
         
         \makecell[l]{SAC \cite{p42}} &  \makecell[l]{Both} &  \makecell[l]{stronger exploratory capabilities, more robust and \\able to adjust more easily in the face of disturbances, faster training} &  \makecell[l]{policy parameter dependency of the distribution} \\ \hline
         
         \makecell[l]{QMIX \cite{r41}} &  \makecell[l]{Value-based} &  \makecell[l]{richer action-value functions that can be decentralized\\ and trained in a centralized end-to-end manner} &  \makecell[l]{it has not been demonstrated to scale to large numbers\\ of agents approximate total Q value in the form of a sum \\of all agent's action values Q} \\ \hline
         
         \makecell[l]{Meta Q learning \cite{r37}} &  \makecell[l]{Value-based} & \makecell[l]{transferring and reusing learned experience} & \makecell[l]{lack of performance analysis on different tasks or environments\\ lack of analysis of the transferred knowledge} \\ \hline

        \makecell[l]{PPO \cite{r50}} & \makecell[l]{Policy-based} & \makecell[l]{obtains better performance than TRPO in\\ robot manipulation in virtual simulation environment}  & \makecell[l]{only data from the previous round of strategies\\ were used not data from all past strategies} \\ \hline
        
        \makecell[l]{ERL \cite{r42}} & \makecell[l]{Policy-based} & \makecell[l]{noticeably better performance than earlier DRL and EA techniques} & \makecell[l]{strong basin of attraction of the local minima \\confines ERL in the early stage} \\ \hline
        
       \makecell[l]{TD3 \cite{r43}} & \makecell[l]{Both} & \makecell[l]{TD3 significantly improves  performance and learning  speed of DDPG} & \makecell[l]{exploration is inefficient and difficult to be widely adopted \\when Critic estimation and true Q are far apart} \\ \hline
       
    \end{tabular}
    }
\end{table*}

 \begin{table*}[!ht] 
    \centering
    \caption{Reinforcement Learning Algorithms compare - Model based}
    \label{Reinforcement Learning Algorithms compare - Model based}
    
    \resizebox{\textwidth}{0.05\textheight}{\begin{tabular}{|l|l|l|}
    \hline
        \makecell[l]{Method} & Advantages & Disadvantages \\ \hline
        
        \makecell[l]{GPS\cite{P37} } & \makecell[l]{effectively balances the search from a good local optimum} &  \makecell[l]{limited by the low efficiency of conventional control method } \\ \hline

        \makecell[l]{CEM\cite{r51}} & \makecell[l]{no need to compute gradients, fast convergence, parallel computation } & \makecell[l]{Poor effect with high dimensionality} \\ \hline

        \makecell[l]{I2A\cite{r44} } & \makecell[l]{compared to most existing model-based RL methods, I2A learns to interpret predictions from models of\\ the learning environment and uses the predictions as additional context for deep policy networks} & \makecell[l]{slow convergence, no significant improvement in data efficiency} \\ \hline

        \makecell[l]{MBMF RL \cite{r52} } & \makecell[l]{can be trained only once to apply the model to different tasks by modifying the reward function,\\ without retraining the intelligent body model} & \makecell[l]{lack of performance analysis on model-free learners to provide \\further sample efficiency gains} \\ \hline

        \makecell[l]{MBVE\cite{r46}} & \makecell[l]{the prediction of target Q-value combines short-term estimation based on environmental models \\and long-term prediction based on target networks, resulting in more accurate prediction results } & \makecell[l]{lack of performance analysis on model-free sample complexity reduction. } \\ \hline

    \end{tabular}}
\end{table*}

\subsubsection{Reinforcement Learning conclusion}
One of the main advantages of Reinforcement Learning over traditional machine learning is that it can actively learn to get the feedback it needs from the environment. Another advantage is that Reinforcement Learning algorithms learn strategies that can be executed in dynamic environments, which is closer to what is commonly understood as artificial intelligence than supervised and unsupervised learning. Reinforcement Learning enables agents to better understand their environment and learn higher-level causality.  

Furthermore, with the combination of deep learning and RL, DRL can build independent dynamic model that well adapt to the environment, such as perform tasks in continuous state space or with high-dimensional pixel inputs. While for conventional RL it is more suitable to deal with finite state space \cite{robotcontrol}. Based on the deep neural network, DRL can extract low-dimensional features from high-dimensional input data like image or pixel.  

From previous studies, several challenges regarding RL remain to be addressed. Traditional RL is limited by memory complexity, computational complexity and sample complexity which partially relieved by combining with deep learning algorithm \cite{P37}. However, the optimal policy still needs to be obtained by trials and errors with the environment and learner can only receive reward as feedback reference \cite{P37}. In addition, DRL sampling efficiency is a main concern and it's hard to manually define the optimal or relatively correct reward function which is the foundation of RL. Apart from above, in RL, the agent needs to deal with long-term dependency, credit assignment problem that the reward signal may be delay due the strong temporal correlation between learner observations and actions \cite{P37}. 
 

\subsection{Imitation Learning}
Robots can now acquire and master manipulation abilities in dynamic contexts with machine learning, which makes up for the drawbacks of conventional artificial intelligence methods. Imitation Learning is a type of robot learning in which a robot learns an expert's performance to perform a specific task \cite{P34}. The main goal of IL is to learn an expert's behavior through imitation, meaning that the robot can replicate the behavior and apply it to a new scenario \cite{P33}. 

By comparing with traditional robot control methods, imitation learning has the following advantages \cite{P33}. 

\begin{itemize} 
\item[$\bullet$]  
Better learning efficiency. The main advantage of IL is its high learning efficiency. It also allows faster and wider dissemination of useful behaviors, increasing the robot's adaptability and efficiency. 
\end{itemize} 

\begin{itemize} 
\item[$\bullet$]  
More adaptable. The robot can observe the actions of others and imitate them and can quickly learn the actions and apply them to new environments. 
\end{itemize} 

\begin{itemize} 
\item[$\bullet$]  
Better communication efficiency. IL offers the possibility of nonverbal communication, allowing the robot to learn autonomously from experts' supervision. 
\end{itemize} 
Based on these advantages, IL has developed rapidly in recent research. The process of IL for robot manipulation consists of the following basic stages: demonstration handling, feature description and motion learning \cite{P33, r47}. This section will review the current state of IL research based on the above process.

\begin{figure}[ht]  
\centering 
    \includegraphics[width=\columnwidth]{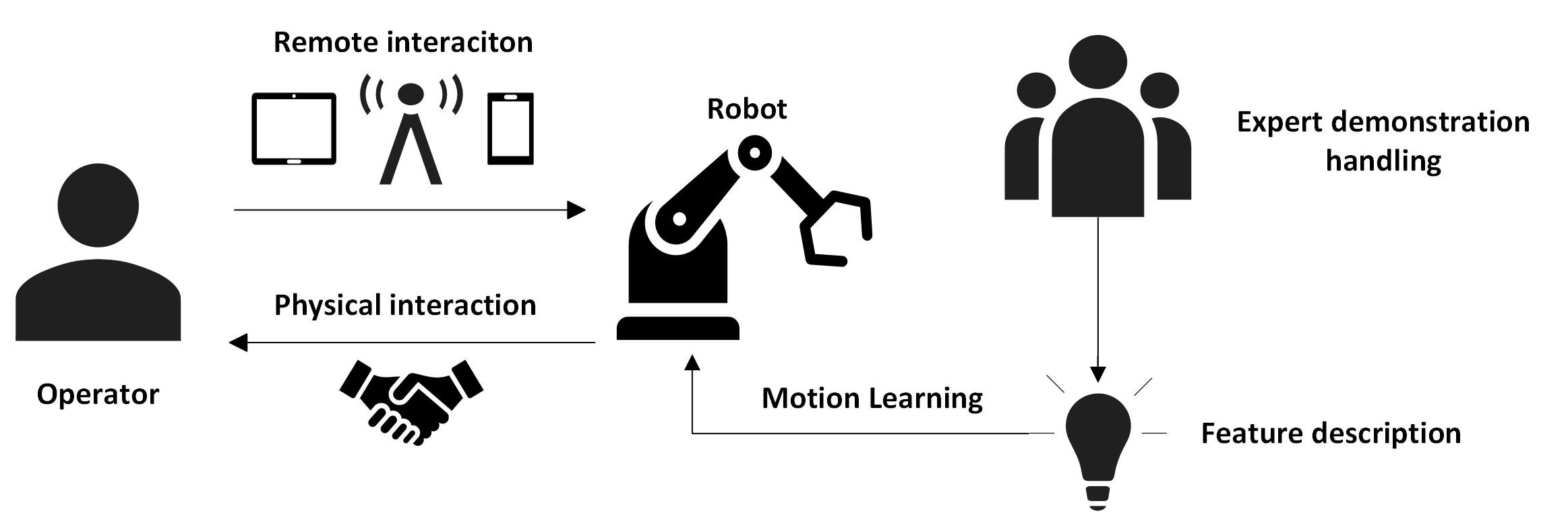} 
    \caption{Imitation Learning Process} 
    \label{Fig: Imitation Learning Process} 
\end{figure}

\subsubsection{Demonstration Handling} 
In IL, observing the expert's demonstration behavior is the basis of the IL process \cite{P33Attia2018}. IL does not receive the same rewarding feedback on the task as RL. IL can use the expert's demonstration of the task to learn policy and to reproduce similar action behaviors \cite{P34}. Previous studies have shown that IL is mainly handling two types of demonstrations which are direct demonstration and indirect demonstration \cite{P33}. 

\paragraph{Direct demonstration}
Direct demonstration means that the demonstration samples are obtained directly through the robot. Direct demonstration makes the process simple, straightforward and accurate. The direct demonstration can be divided into kinesthetic demonstration and remote operation demonstration.   

Kinesthetic teaching involves the operator directly engaging with the robot and guiding it to complete a specific task. In this process, the robot autonomously collects valid information. The goal of kinesthetic teaching is to physically guide the robot to exhibit the desired behavior \cite{P35}.  However, kinesthetic training operations are limited to multi-degree-of-freedom robots and do not perform well in multi-degree-of-freedom robot control. 

Remote manipulation demonstrations can be performed remotely with a joystick or control panel, wearable devices, haptic sensors, etc. The demonstrator is not limited to physical contact with the robot but can perform the demonstration from anywhere, which also ensures the safety of the operator. However, this demonstration method has some limitations like delay and noise of remote data transmission, and the accuracy of teaching cannot be guaranteed by the equipment. In addition, the current tele-demonstration equipment can only teach based on trajectory and posture, which lacks sufficient information.  

\paragraph{Indirect demonstration}  
Indirect demonstrations are constructed in the environment and do not involve contact movement of the robot, which allows the robot to gather operational information during this teaching process. Indirect demonstrations usually involve collecting information about the expert's movements through vision systems \cite{P33.Sermanet} or wearable devices \cite{P33.Edmonds}. Demonstration of visual information can be achieved by looking at pictures, but it requires a high learning rate and lacks sufficient tactile information. Wearable devices can collect information about the demonstrator's movements and tactile strength, but the accuracy and noise immunity of the information needs to be improved due to the limitations of the device data transmission.  

\subsubsection{Feature Description} 
After the expert demonstrations are collected, useful features need to be extracted and applied to the learning process. This is because the robot operation is complex and noisy, and there is a lot of useless and redundant information in the demonstration data. Robot control demonstration in IL can be divided into three categories: symbolic feature, trajectory feature, and action-state spatial feature \cite{P33}. 

\paragraph{Symbolic feature} 
The symbolic feature means that the robot learns to generate a series of options, each of which consists of action sequences ordered by time. Symbolic representation allows the same action to be used repeatedly for different tasks. It is a practical and workable method for enhancing the intelligence of multi-modal integrated robots and offers a workable answer to challenging, multi-step task learning issues \cite{P33}. 

\paragraph{Trajectory feature} 
The trajectory feature is a lower-level representation compared with symbolic representation, which can be understood as mapping the task-related conditions and trajectories \cite{P33}. The task-related trajectory conditions can be the gripper's initial location and target location. Hence, the movement of the robot arm can be represented as system states in time series. This representation method needs as much dynamic information as possible and this could lead to poor performance on image-based and haptic-based features \cite{P33}.  

\paragraph{Action-state spatial feature}
Compared with the two representation methods above, action-state spatial feature methods bring a series of action-state pairs decided based on a specific state, rather than bringing numerous strategies options to choose. By mapping the task state conditions to the manipulation actions, this method is widely used in the stability observation of dynamic system behavior cloning \cite{P33.Khansari2011}. This representation method suffers from the short-term behavior action-state space which will lead to accumulation error in long-term representation \cite{P33}. Features representation can also be categorized into raw features, manually designed features, and extracted features \cite{r47}. This feature classification method is mainly based on different feature sources.  

\subsubsection{Imitation Learning algorithms} 
Imitation Learning aims to train the robot until it learns the manipulation through experts' demonstration, also with the ability to implement skills in unprecedented environments. Most studies viewed robot manipulation tasks as Markov Decision Process (MDP) which can be described as $M\left\{ S, A, P, r, \gamma\right\}$. MDP is generally used in Imitation Learning algorithms as well. The paragraphs below will review different types of IL algorithms from three main methods: Behavior Cloning, Inverse Reinforcement Learning, and Generative Adversarial Imitation Learning.

\paragraph{Behavior Cloning}
Behavior Cloning is a direct learning policy, which allows the distribution of agent-generated state-action trajectories to match the given demonstration trajectory \cite{P11.72}. 
$s_t$ of $(s_t, a_t)$ in the expert data is considered as the sample input and $a_t$ as the label, and the learning objective is
\begin{align}
    \theta^* = \arg\!\min\limits_\theta E_{(s,a)\sim D} \left[ L(\pi_\theta (s), a) \right]
\end{align}
where $D$ is the expert's data set and $L$ is the loss function in the corresponding supervised learning framework. 
A limitation of BC is that it’s hard to adapt to environmental changes by replicating the same learned actions. Robot manipulation basic units are identified by machine learning methods with the development of statistical learning. Emerging Behavioral Cloning algorithms can be classified as model-based BC and model-free BC based on if the model/environment distribution is known. To briefly explain model-based and model-free, model-based approaches are defined as whether they learned dynamic models during the Imitation Learning process \cite{P34}. So, model-free approaches indicate they have no model-learning step during the imitation policy learning process \cite{P34}. On the other hand, BC can also be classified based on demonstration feature descriptions such as symbolic representation BC, trajectory representation BC and action-state spatial representation BC. BC classification methods can be freely combined based on the above two categories, and the same applies to the classification of IL.  

This section summarizes some widely used BC algorithms that have been used in previous studies. Hidden Markov Model (HMM) were proposed by Takeda and trained a robot control strategy to dance with human, but this model is only able to deal with discrete states and state transitions that cannot implement on continuous robot motion control tasks \cite{P11.73}. The development of Gaussian Mixture Model (GMM) and Gaussian Mixture Regression (GMR) solve the problems above and are used to describe different trajectory stages' uncertainty \cite{robotcontrol}.  

In [48] also proposed an evolutional BC algorithm called Behavioral Cloning from Observation (BCO), which is an Imitation Learning algorithm without the need for access to demonstrator actions and post-demonstration environment interaction. BCO has been proven that performance improves with more environmental interactions to better estimate experts’ actions \cite{r48}.  

Furthermore, dynamic motion primitives and mixed motion primitive methods can generate trajectory representations that are continuous, stable and generalizable. The Imitation Learning methods above are learning the demonstration information to generate actions and trajectories, which are not executed during robot processing \cite{P33}. To collect high-quality demonstrations, a VR tele-operation system proposed with visuomotor learning (VL) control strategy \cite{P11.77}.  

However, most of BC algorithms are limited by the need for sufficient samples, which means that BC cannot learn strategies that are not present in the sample. 

\paragraph{Inverse Reinforcement Learning}
Inverse Reinforcement Learning (IRL) can solve the insufficient sample problems of Behavior Cloning and reward function definition problem of Reinforcement Learning. The Inverse Reinforcement Learning method evaluates actions performance by reward function which summarizes the behavior description. The reward function of IRL assumes the optimal of expert demonstrations and shows the expert's purpose \cite{P33}.  

IRL is more adaptable in different environments compared with BC. There are many studies of IRL that showed that it is more task-related. And the proper policies are obtained from the previous reward function after receiving new information from the environment and model. IRL could result in a generalized strategy even with insufficient demonstration samples \cite{P33.Justin}.  
 
Some IRL algorithms will be mentioned in this paragraph. The first one is the IRL of maximum entropy model that resolves the reward function random deviation caused by expert teaching preferences \cite{P33}, while this method needs to know the system state transition probability \cite{robotcontrol}. 
In previous research on maximum entropy IRL method by Ziebart \cite{IRL.Adams}, he took a probabilistic analysis of any observed deterministic behavior single trajectory with the probability as below:
\begin{align}
    P(\tau_m | \theta) = \frac{1}{Z(\theta)} e^{\theta^{T}f_{\tau_m}}
\end{align}\\
that $\theta$ is the feature weights vector, $f_{\tau_m} = \sum\limits_{s_j \in \tau_m}f_{s_j}$ is the count of path feature, and $Z(\theta)$ is the partition function with feature weights provided. And non-deterministic behavior of observing any single trajectory is:
\begin{align}
    P(\tau|\theta,P) \approx \frac{e^{\theta^{T}f_Tau}}{Z(\theta, P]} \prod_{s_{t+1}, a_t, s_t} P_{s_t, s_{t+1}} ^a
\end{align}
Then with maximizing the log-likelihood $L(\theta)$ of the entropy distribution below to find reward weights \cite{IRL.Adams}. 
\begin{align}
    \theta^* = \arg\!\max_\theta \sum\limits_{m=1} ^M \log P(\tilde{\tau_m}| \theta, P)
\end{align}
which $\tilde{\tau_m}$ represents the $m^{th}$ observed trajectory. Then using gradient descent to max the likelihood:
\begin{align}
    \nabla{L(\theta)} & = \bar{f} - \sum\limits_\tau P(\tau|\theta,P) f_\tau \\
    & = \tilde{f} - \sum\limits_{s_i} D_{s_i} f_{s_i}
\end{align}
and $\tilde{f}$ is the count of empirical features, $D_{s_i}$ is the frequency of the expected visits \cite{IRL.Adams}. 

Apart from this, the maximum margin principle maximized the difference between optimal policies and other policies \cite{P11.81}. The maximum relative entropy model is proposed by Boularias to resolve the model-free problem \cite{P11.89}. Recently, DeepMimic algorithm is introduced with neural network-modelled policies and training with proximal policy optimization algorithm in \cite{P9.18}. In \cite{P9.18}, DeepMimic methods with imitation reward and task-specific reward synthesize the controller to imitate the reference actions and satisfy the task objectives. Additionally, Trajectory-ranked Reward Extrapolation (T-REX) is proposed in \cite{r48} achieved improved performance on Atari and MuJoCo benchmark tasks compared with BCO and GAIL methods.  

Inverse reinforcement learning recovers the reward function based on the expert demonstration optimality, but it is less effective in exchanging information with expert to improve the policy further \cite{robotcontrol}. Also, IRL has an expensive computation cost that it requires Reinforcement Learning in inner loop calculation \cite{GAIL}. 

\paragraph{Generative Adversarial Imitation Learning}
The Generative Adversarial Imitation Learning (GAIL) method compares the generated policy and expert policy differences in discriminator\cite{robotcontrol}. This method defines expert behavior by using generative adversarial training to fit distributions of states and actions \cite{GAIL}. 

GAIL algorithm has a discriminator and a strategy. The strategy $\pi$ is the equivalent of a generator in a generative adversarial network, where given a state, the strategy outputs the action that should be taken in that state, while the discriminator $D$ takes as input a state action pair $(s,a)$ and outputs a real number between 0 and 1 a real number indicating the probability that the discriminator considers the state action pair $(s,a)$ to be from an intelligence policy rather than an expert. The goal of the discriminator is to keep the output of the expert data as close to 0 and the output of the imitator strategy as close to 1 as possible, so that the two sets of data can be distinguished \cite{GAIL.Ho2016}. Thus, the loss function of the discriminator $D$ is:
\begin{align}
    L(\phi) = -E_{\rho_\phi} \left[ \log D_\phi (s,a) \right] - E_{\phi_E} \left[ \log (1-D_\phi (s,a))\right]
\end{align}
which $\phi$ is the parameter of the discriminator $D$. 

With a discriminator $D$, the goal is to produce trajectories that can be recognized as expert trajectories by the discriminator. As the adversarial process continues, the data distribution generated by the imitator's strategy will approach the real expert data distribution for the purpose of imitation learning.
 
There is also a model-based GAIL proposed by Baram in \cite{P11.94} used forward model and discriminator gradients to train policies \cite{P11.94}. This model-based GAIL shows an IL method for hierarchical optional framework strategy \cite{P33}. In \cite{P34}, the author also proposed an Imitation Learning from observation(IfO) model-free adversarial methods. This imitation strategy feeds individual states into a discriminator that is trained to distinguish between data from the imitator and data from the demonstrator. The data is collected during strategy execution \cite{P34}. There is another IfO adversarial approach in \cite{P34.Stadie2017} focused on different viewpoints between robot (imitator) and expert (demonstrator). A new classifier is proposed to use discriminator early layer outputs as input and try to distinguish different viewpoints' data \cite{P34.Stadie2017}.  

GAIL introduces generative adversarial networks for IL that lead to better high-dimensional situation performance compared with BC and IRL. Apart from this, GAIL is also very sample efficient in expert data \cite{GAIL}. However, model-free GAIL method is not quite sample efficient during training environment interaction \cite{GAIL}.

\begin{table*}[h] \scriptsize
    \centering
    \caption{Imitation Learning Algorithms compare}
    \label{Imitation Learning Algorithms compare}
    \resizebox{\textwidth}{0.15\textheight}{\begin{tabular}{|c|c|c|c|}
    \hline 
    \makecell[l]{Method} & Category & Advantages & Disadvantages \\
    \hline 
    \makecell[l]{HMM\cite{P11.73} } & \makecell[l]{BC} & \makecell[l]{widely used in the fields of language recognition, natural \\language processing, pattern recognition, etc.} & \makecell[l]{HMM depends only on each state and\\ its corresponding observer }\\ \hline

    \makecell[l]{GMM \cite{r53}]} & \makecell[l]{BC} & \makecell[l]{solve the continuous robot motion control tasks problems\\ and can be used to describe different trajectory stages' uncertainty} & \makecell[l]{may fall into local extremes, which is very relevant\\ to the selection of the initial value}\\ \hline

    \makecell[l]{BCO \cite{r56} } & \makecell[l]{BC} & \makecell[l]{performance improves with more environmental interactions\\ to better estimate experts’ actions} & \makecell[l]{lack of performance analysis on devise \\tasks and with hyperparameter dependency}\\ \hline

    \makecell[l]{max-entropy IRL \cite{P33}} & \makecell[l]{IRL} & \makecell[l]{resolves the reward function random deviation \\caused by expert teaching preference} & \makecell[l]{system state transition probability dependency }\\ \hline

    \makecell[l]{DeepMimic \cite{P9.18}} & \makecell[l]{IRL} & \makecell[l]{with imitation reward and task-specific reward\\ synthesize the controlled to imitate the reference\\ actions and satisfy the task objectives} & \makecell[l]{the reference clips are highly structured\\ data synthesized by motion capture, and the data\\ itself requires a high cost to acquire}\\ \hline

    \makecell[l]{f-IRL \cite{r55}} & \makecell[l]{IRL} & \makecell[l]{pick up on behaviors from a manually created\\ target state density or implicitly from professional observations} & \makecell[l]{high computation cost on IRL process}\\ \hline

    \makecell[l]{T-REX \cite{r48}} & \makecell[l]{IRL} & \makecell[l]{achieved improved performance on Atari and MuJoCo benchmark\\ tasks compared with BCO and GAIL methods} & \makecell[l]{relatively high computation and time cost}\\ \hline

     \makecell[l]{VAE\cite{r57}} & \makecell[l]{GAIL} & \makecell[l]{can unsupervised infer the hidden variables in the sample} & \makecell[l]{modal inference ambiguity}\\ \hline

     \makecell[l]{IfO \cite{P34}} & \makecell[l]{GAIL} & \makecell[l]{possible to determine the position and orientation \\of the objects in a cluttered video} & \makecell[l]{lack of performance analysis on physical robots}\\ \hline

     \makecell[l]{MGAIL \cite{r58}} & \makecell[l]{GAIL} & \makecell[l]{high sample utilization to alleviate high variance problems} & \makecell[l]{extending the application of imitation learning to\\the real problem of high cost of interaction\\ between agent and its environment}\\ \hline

    \makecell[l]{TPIL \cite{r58}} & \makecell[l]{GAIL} & \makecell[l]{can make observation uncorrelated with observation perspective} & \makecell[l]{mode collapse problem does not resolve}\\
    \hline 
    
    \end{tabular}}
\end{table*}


\section{Challenges and Future Works}
In the respect of human-robot interaction control, the most popular and widely used robot manipulation method currently is Reinforcement Learning. However, most of the existing research on RL-based HRI is limited in simulation environments due to the potential damage to robots and injury to humans when collecting and training data. Therefore, the application capability of the system in the real world will be considered, so that the performance analysis of HRI is not limited to the simulated environment.  

Additionally, RL control relies on the reward function that interacts with the environment to find the optimal policy or action. However, the choice of the reward function needs to be set manually, and small changes in the reward function can result in very different optimal solutions. In addition, in real-world scenarios where the reward function is difficult to determine or the reward signal is sparse, a randomly designed reward function cannot guarantee that the trained strategies will meet the needs of RL. Although DRL solves the problem of processing high-dimensional raw signals, the slow convergence and long computation time lead to high consumption of training data of RL. 

Behavioral Cloning accomplishes tasks by directly imitating expert behavior. Inverse Reinforcement Learning is a non-single-step decision-making algorithm for sequential decision-making. The reward function is learned through expert sampling to solve the problem that it is difficult to quantify the setting of the reward function and the reward function in RL. But these two IL methods also have some disadvantages. 

For example, BC has been restrained when the amount of data is relatively small. At the same time, BC can only learn the provided samples. It doesn't have the ability to generalize expert behavior, because BC can only make relatively accurate predictions under the state distribution of expert data. In addition to this, BC also suffers from error-compounding problems. Error compounding is because in the process of robot operation if there is a little error, it may lead to the next encountered state that has not been seen in the expert data. At this time, since it has not been trained in this state or a relatively similar state, the strategy may randomly select an action, which will cause the next state to further deviate from the data distribution of the expert strategy. 

In order to solve the problem of error compounding, Stéphane Ross et al. proposed the DAgger (Dataset Aggregation) algorithm. This is an algorithm based on online learning. The Dagger algorithm is the introduction of online iteration in BC. But the DAgger algorithm also has obvious flaws. Due to the mechanism of online learning, it requires a long and continuous request for experts to supplement the teaching mark. This brings a great workload and difficulties to the application of such algorithms in real environments. And there is no essential difference between the DAgger algorithm and the general BC algorithm, and they have never been separated from the use of samples as a supervision signal for imitation learning, so it is always limited by the need for numerous samples to achieve a good supervision effect and does not try to learn rules from samples as new supervision signals. 

BC is struggled to effectively learn behavior rules through expert samples. Another IL method, Inverse Reinforcement Learning, learns reward functions through expert samples to improve its own strategy level. The proposal of IRL provides a solution to the problem that the RL reward function is difficult to set. However, IRL is limited by the problem of low efficiency, high computational complexity and accurate inference of reward functions and policies. IRL has a large amount of calculation, which is equivalent to running the RL algorithm once in each inner loop. IRL also suffers from obstacles to accurately reasoning about reward functions or optimal expert policies. Because IRL first learns the reward function based on expert samples, and then improves its own strategy level by applying the reward function in the RL algorithm. But what is found is not necessarily the true reward function. Because based on limited sampling, many reward functions can explain the data provided or observed by experts, and there is no guarantee that policies with improved reward functions can perform well in different environments \cite{IRL.Arora}. 

GAIL uses the framework of generative adversarial networks for imitative learning to overcome the shortcomings of BC, IRL, and it can perform well in large-scale problems. Based on the generative adversarial network framework, GAIL's policy and reward function models can use neural networks to automatically extract the abstract features of samples. As a result, GAIL has stronger representational capabilities. Moreover, GAIL uses strategy as the learning target directly, it uses an efficient strategy gradient method to train the strategy model. Thus, GAIL can avoid the internal computation process that requires large computational resources for other IL methods and has more efficient computational power. Work has shown that GAIL can perform well in complex large-scale problems such as autonomous driving, simulation, and realistic robot manipulation \cite{r15}.  

However, GAIL still faces many bottlenecks. Among them, the mode collapse problem and the inefficient use of environmental interaction samples are particularly important. The mode collapse problem originates from Generative Adversarial Nets (GANs), which leads to the loss of diversity in the samples generated by GAIL. The problem of inefficient sample generation stems from the assumption of stochastic policy and model-free policy learning, which will make GAIL unsuitable for practical applications with high sample acquisition costs. To address the mode collapse problem, GAIL is proposed to be improved by using a variant form of GANs. The improvement methods include multimodal hypothesis-based improvement, generative model-based improvement, etc. To address the problem of inefficient utilization of generated samples, GAIL is proposed to be improved by using RL techniques. The improvement methods include improvement based on a dynamic model, improvement based on deterministic strategy, improvement based on the Bayesian method, etc. \cite{r15}. 

\section{Conclusion}
In this paper, robot manipulation algorithms for Human-Robot Interaction systems have been reviewed. This review paper has demonstrated growing interest and pertinent technological advancements in this area, spurring additional research and development in the area of HRI. With the overview of different robot manipulation elements in HRI field, this paper explained the formulation of robot manipulation in Markov Decision Processes. Then we discussed different robot manipulation methods aiming at learning skill policies mainly including Reinforcement Learning and Imitation Learning. Among the manipulation methods in the literature, we also review the current challenges and state-of-art solution research in Section 4. Finally, the results show the enthusiasm and progress of current research and development prospects of robotic manipulation in the field of HRI. It also provides strong technical support for the development of HRI systems.


%




\ifCLASSOPTIONcaptionsoff
  \newpage
\fi



\bibliographystyle{IEEEtran}
\bibliography{ref}

\end{document}